\documentclass[10pt,twocolumn,letterpaper]{article}

\usepackage[pagenumbers]{cvpr} 

\definecolor{cvprblue}{rgb}{0.21,0.49,0.74}
\usepackage[pagebackref,breaklinks,colorlinks,allcolors=cvprblue]{hyperref}

\usepackage{multirow}
\usepackage{graphicx}
\usepackage[para]{footmisc}
\usepackage{balance}
\usepackage{multibib}
\usepackage{soul}
\usepackage{booktabs}
\usepackage{enumerate}
\usepackage{enumitem}
\usepackage{wrapfig}

\def\paperID{****} 
\def\confName{CVPR}
\def\confYear{2026}

\title{DMin: Scalable Training Data Influence Estimation for Diffusion Models}

\author{
Huawei Lin $^{1}$\hspace{.2in}
Yingjie Lao $^{2}$\hspace{.2in}
Weijie Zhao $^{1}$\\
$^1$\ Rochester Institute of Technology\hspace{.2in}
$^2$\ Tufts University\\
\texttt{hl3352@rit.edu}\hspace{.2in}
\texttt{Yingjie.Lao@tufts.edu}\hspace{.2in}
\texttt{wjz@cs.rit.edu}
}

\begin{document}

\twocolumn[{
\vspace{-.1in}
    \renewcommand\twocolumn[1][]{#1}%
    \maketitle
    \begin{center}
        \vspace{-0.25in}
        \includegraphics[width=1\linewidth]{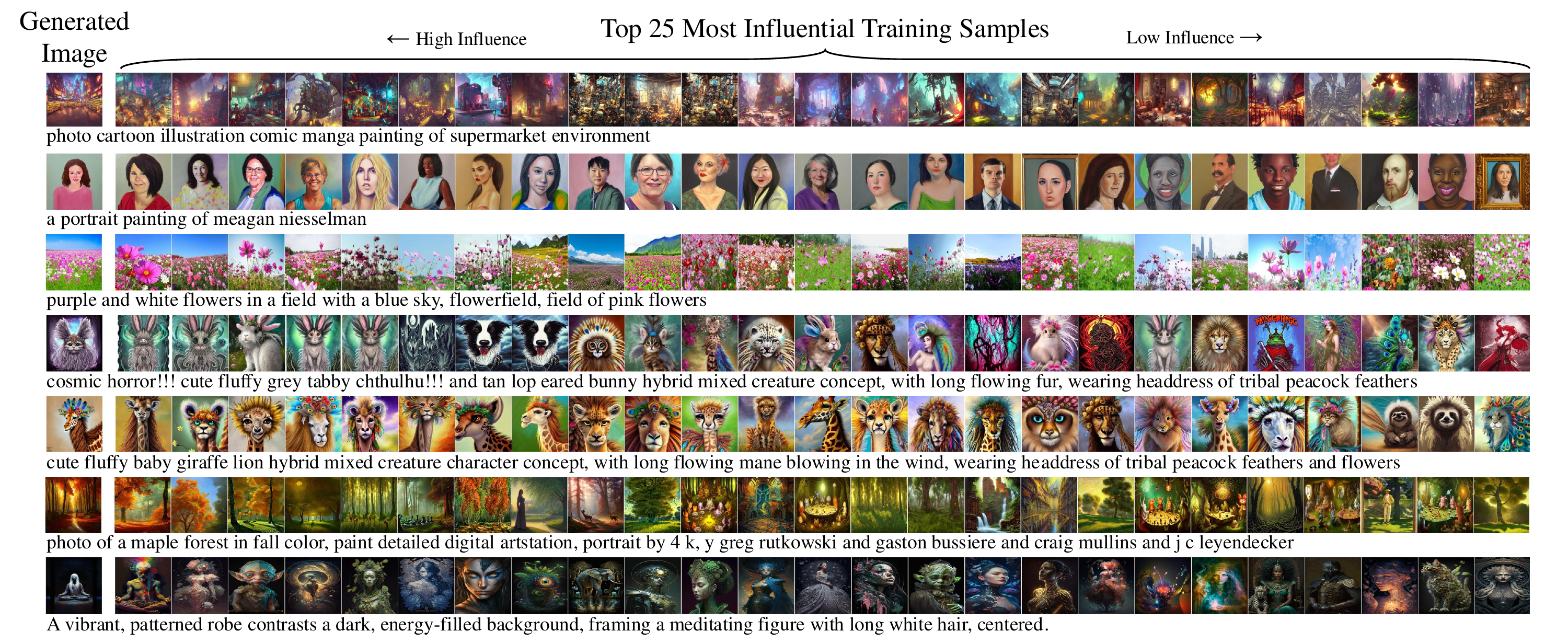}
        \vspace{-0.25in}
        \captionof{figure}{Examples of influential training samples, with prompts displayed below generated image. (SD 3 Medium with LoRA, $v=2^{16}$).}
        \label{fig:top_figure}
        \vspace{0.1in}
    \end{center}
}]

\begin{abstract}
Identifying the training data samples that most influence a generated image is a critical task in understanding diffusion models (DMs), yet existing influence estimation methods are constrained to small-scale or LoRA-tuned models due to computational limitations. To address this challenge, we propose \texttt{DMin} (\textbf{\underline{D}}iffusion \textbf{\underline{M}}odel \textbf{\underline{in}}fluence), a scalable framework for estimating the influence of each training data sample on a given generated image. To the best of our knowledge, it is the first method capable of influence estimation for DMs with billions of parameters. Leveraging efficient gradient compression, \texttt{DMin} reduces storage requirements from hundreds of TBs to mere MBs or even KBs, and retrieves the top-$k$ most influential training samples in under 1 second, all while maintaining performance. Our empirical results demonstrate that \texttt{DMin} is both effective in identifying influential training samples and efficient in terms of computational and storage requirements.
\end{abstract}

\vspace{-.1in}
\section{Introduction}

Diffusion models have emerged as powerful generative models, capable of producing high-quality images and media across various applications~\cite{DBLP:journals/pami/CroitoruHIS23, DBLP:journals/csur/YangZSHXZZCY24, DBLP:journals/corr/abs-2208-11970, DBLP:conf/iccv/ZhangRA23, DBLP:journals/corr/abs-2505-13439}. Despite their impressive performance, the datasets used for training are often sourced broadly from the internet~\cite{DBLP:conf/nips/SchuhmannBVGWCC22, DBLP:conf/acl/WangMMYHC23, DBLP:conf/sigir/Srinivasan0CBN21, DBLP:conf/naacl/PanLRCYZZX25, DBLP:journals/corr/abs-2502-13141}. This vast dataset diversity allows diffusion models to generate an extensive range of content, enhancing their versatility and adaptability across multiple domains~\cite{DBLP:conf/nips/LiWZULYNPG23, DBLP:conf/iccv/ChenSSL23}. However, it also means that these models may inadvertently generate unexpected or even harmful content, reflecting biases or inaccuracies present in the training data.

This raises an important question: \textit{given a generated image, can we estimate the influence of each training data sample on this image?} Such an estimation is crucial for various applications, such as understanding potential biases~\cite{DBLP:conf/iclr/KongSH22, DBLP:journals/corr/abs-2312-05586} and improving model transparency by tracing the origins of specific generated outputs~\cite{DBLP:conf/icml/KohL17, DBLP:journals/corr/abs-2405-13954, DBLP:journals/corr/abs-2308-03296}.

Recently, many studies have explored influence estimation in diffusion models~\cite{DBLP:journals/corr/abs-2410-13850, DBLP:conf/iclr/KwonWW024, DBLP:journals/corr/abs-2410-13850, DBLP:conf/emnlp/OguejiAOGHK22, DBLP:journals/corr/abs-2312-06205}. These methods assign an influence score to each training data sample relative to a generated image, quantifying the extent to which each sample impacts the generation process. For instance, DataInf~\cite{DBLP:conf/iclr/KwonWW024} and K-FAC~\cite{DBLP:journals/corr/abs-2410-13850} are influence approximation techniques tailored for diffusion models. However, they are both second-order methods that require the inversion of the Hessian matrix. To approximate this inversion, they must load all the gradients of training data samples across several predefined timesteps. Notably, in the case of the full-precision Stable Diffusion 3 Medium model~\cite{DBLP:conf/icml/EsserKBEMSLLSBP24}, the gradient of the entire model requires approximately 8 GB of storage. Collecting gradients for one training sample over 10 timesteps would consume $8 \times 10 = 80$ GB. Scaling this requirement to a training dataset of $10,000$ samples results in a storage demand of around 800 TB -- far exceeding the capacity of typical memory or even hard drives. Given that diffusion models are often trained on datasets with millions of samples, this storage demand becomes impractical. Consequently, these methods are limited to LoRA-tuned models or small diffusion models~\cite{DBLP:conf/nips/HoJA20, DBLP:conf/cvpr/RombachBLEO22}. Although some prior works have applied gradient compression, such as SVD~\cite{DBLP:journals/corr/abs-2308-03296} and quantization~\cite{DBLP:journals/corr/abs-2410-13850}, the achieved compression rates are insufficient to maintain performance at this scale.

Alternatively, Journey-TRAK~\cite{DBLP:journals/corr/abs-2312-06205} and D-TRAK~\cite{DBLP:conf/emnlp/OguejiAOGHK22} are first-order methods for influence estimation on diffusion models, which are extended from TRAK~\cite{DBLP:conf/icml/ParkGILM23} on deep learning models. Both approaches utilize random projection to reduce the dimensionality of gradients. However, for large diffusion models, such as the full-precision Stable Diffusion 3 Medium model, the gradient dimensionality exceeds 2 billion parameters. Using the suggested projection dimension of $32,768$ in D-TRAK, storing such a $2\text{B} \times 32,768$ projection matrix requires more than 238 TB of storage. Even if the projection matrix is dynamically generated during computation, the scale of these operations substantially slows down the overall process. As a result, they are only feasible for small models or adapter-tuned models.

\textbf{Challenges.} Although these approaches have demonstrated superior performance on certain diffusion models, several key challenges remain: \textbf{(1) Scalability on Model Size:} Existing methods either require computing second-order Hessian inversion or handling a massive projection matrix, both of which restrict their applicability to large diffusion models. \textbf{(2) Scalability on Dataset Size:} Diffusion models frequently rely on datasets containing millions of samples, making the computation of a Hessian inversion for the entire training dataset impractical. Additionally, storing the full gradients for all training data samples presents a significant challenge. \textbf{(3) 
Fragility of Influence Estimation:} Previous studies have demonstrated the fragility of influence estimation in extremely deep models~\cite{DBLP:conf/acl/LinLX024, DBLP:conf/iclr/BasuPF21, DBLP:journals/nn/EpifanoRMR23, DBLP:conf/aaai/GhorbaniAZ19}. Similarly, we observed this fragility in large diffusion models, regardless of whether they use U-Net or transformer.

To address these challenges, in this paper, we propose \texttt{DMin}, a scalable influence estimation framework for diffusion models. Unlike existing approaches that are limited to small models or LoRA-tuned models, the proposed \texttt{DMin} scales effectively to larger diffusion models with billions of parameters. For each data sample, \texttt{DMin} first computes and collects gradients at each timestep, then compresses these gradients to MBs or KBs while maintaining performance. Following this compression, \texttt{DMin} can accurately estimate the influence of each training data sample on a given generated image or retrieve the top-$k$ most influential samples on-the-fly using K-nearest neighbors (KNN) search, enabling further speedup based on the specific task.

\textbf{Contributions.} The main contributions of this paper are:
\begin{itemize}[itemsep=0pt, topsep=0pt, leftmargin=*, itemsep=0pt, parsep=0pt, partopsep=0pt]
\item We introduce \texttt{DMin}, a scalable influence estimation framework for DMs, compatible with various architectures, from small models and LoRA-tuned models to large-scale models with billions of parameters.
\item To overcome storage and computational limitations, \texttt{DMin} employs a gradient compression technique, reducing storage from around 40 GB to 80 KB per sample while maintaining accuracy, enabling feasible influence estimation on large models and datasets.
\item \texttt{DMin} utilizes KNN to retrieve the top-$k$ most influential training samples for a generated image on-the-fly.
\item Our experimental results demonstrate \texttt{DMin}'s effectiveness and efficiency in influence estimation.
\item We provide an open-source PyTorch implementation with multiprocessing support\footnote{We will release the GitHub link after acceptance.}.
\end{itemize}


\begin{figure*}[t]
\vspace{-.1in}
    \centering
    \includegraphics[width=\linewidth]{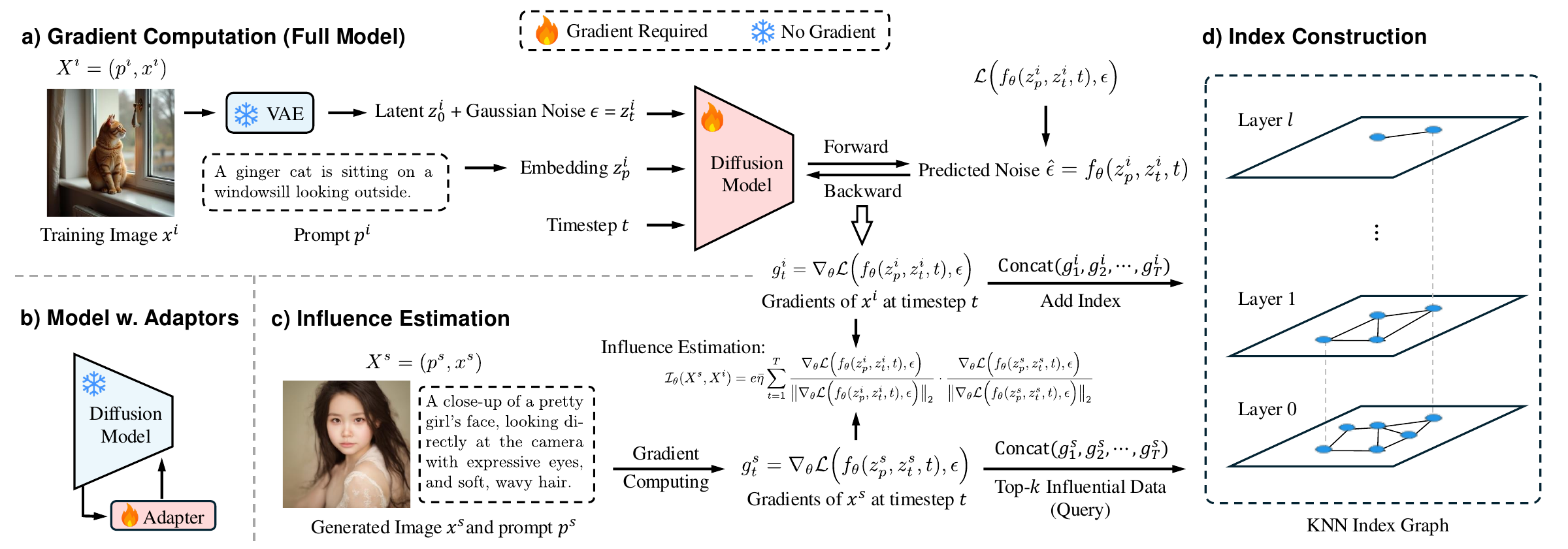}
    \vspace{-.1in}
    \caption{Overview of the proposed \texttt{DMin}.
    \textbf{(a)} In gradient computation, given a training data sample (a pair of prompt $p^i$ and image $x^i$) and a timestep $t$, the data passes through the diffusion model in the same manner as during training. After the backward pass, the gradients $g^i_t$ at timestep $t$ can be obtained.
    \textbf{(b)} For the full model, gradients are collected from the UNet or transformer, whereas for models with adapters, such as LoRA, gradients are collected only from the adapter.
    \textbf{(c)} For a prompt $p^s$ and the corresponding generated image $x^s$, the gradients are obtained in the same way as in Gradient Computation. The influence $\mathcal{I}_\theta(X^s,X^i)$ is then estimated by aggregating gradients across timesteps from $t = 1$ to $T$.
    \textbf{(d)} In some cases, only the most influential data samples are needed; in such instances, KNN can be utilized to retrieve the top-$k$ most influential samples within seconds.
    }
    \label{fig:overview}
    \vspace{-.2in}
\end{figure*}

\section{Influence Estimation for Diffusion Models}
For a latent diffusion model, data $x_0$ is first encoded into a latent representation $z_0$ using an encoder $E$ by $z_0 = E(x_0)$. The model then operates on $z_0$ through a diffusion process to introduce Gaussian noise and iteratively denoise it. The objective is to learn to reconstruct $z_0$ from a noisy latent $z_t$ at any timestep $t \in \{1, 2, \cdots, T\}$ in the diffusion process, where $T$ is the number of diffusion steps. Let $\epsilon_t \sim \mathcal{N}(0, I)$ denote the Gaussian noise added at timestep $t$. We define the training objective at each timestep $t$ as follows:
{\small
\begin{align}
    \theta^*=\arg\min_\theta\mathbb{E}_{z_0, t}\biggl[\mathcal{L}\Bigl(f_{\theta}(z_t, t), \epsilon_t\Bigr)\biggr]
\end{align}
}where $\theta$ represents the model parameters, $z_t$ is the noisy latent representation of $z_0$ at timestep $t$, and $f_{\theta}(z_t, t)$ represents the model's predicted noise at timestep $t$ for the noisy latent $z_t$. $\mathcal{L}(\cdot)$ is the loss function between the predicted noise and the actual Gaussian noise.

Given a test generation $x^s$, where $x^s$ is generated by a well-trained diffusion model with parameters $\theta^*$, the goal of influence estimation is to estimate the influence of each training data sample $x^i$ ($1\le i\le N$) on generating $x^s$, where $N$ is the size of the training dataset. Let $z^i_t$ represent the latent representation of $x^i$ at timestep $t$, and let $z^s_t$ denote the latent representation of the test generation $x^s$ at timestep $t$.

In the $\alpha$-th training iteration, the model parameters $\theta_{\alpha + 1}$ are updated from $\theta_{\alpha}$ by gradient descent on the noise prediction loss for batch {\small$B=(B_z,B_t,B_\epsilon)$}:
{\small
\begin{align}
\label{equ:sgd}
    \theta_{\alpha+1}=\theta_\alpha - \eta_{\alpha}\frac{1}{|B|}\sum_{(z_t, t, \epsilon_t)\in B}\nabla_{\theta_{\alpha}}\mathcal{L}(f_{\theta_{\alpha}}(z_t, t), \epsilon_t)
\end{align}
}where $\eta_\alpha$ denotes the learning rate in the $\alpha$-th iteration, $(z_t^i, t^i, \epsilon_t^i) \in B$, and the contribution of $(z_t^i, t^i, \epsilon_t^i)$ to the batch gradient is {\small $\frac{1}{|B|}\nabla_{\theta_{\alpha}}\mathcal{L}(f_{\theta_{\alpha}}(z^i_t, t^i), \epsilon^i_t)$}. The influence of this training iteration $z^i_t$ with respect to $z^s_t$ on timestep $t$ can be quantified as the change in loss:
{\small
\begin{align}
\label{equ:influence_sub}
    \mathcal{I}_{\theta_{\alpha + 1}, t}(x^s, x^i) = \mathcal{L}\Bigl(f_{\theta_\alpha}(z^s_t, t), \epsilon^i_t\Bigr) - \mathcal{L}\Bigl(f_{\theta_{\alpha + 1}}(z^s_t, t), \epsilon^i_t\Bigr)
\end{align}
}where $z^s_t = E(x^s) + \epsilon^i_t$, denoting the latent representation of $x^s$ after adding Gaussian noise $\epsilon^i_t$, and $\mathcal{I}_{\theta_{\alpha + 1}, t}(x^s, x^i)$ represents the influence of $x^i$ with respect to $x^s$ at the $\alpha$-th iteration with timestep $t$. Then $\mathcal{L}\Bigl(f_{\theta_{\alpha + 1}}(z^s_t, t), \epsilon^i_t\Bigr)$ can be expanded via Taylor expansion:
{\small
\begin{align}
\label{equ:taylor_expansion}
    &\hspace{-.05in}\mathcal{L}\Bigl(f_{\theta_{\alpha + 1}}(z^s_t, t), \epsilon^i_t\Bigr) = \mathcal{L}\Bigl(f_{\theta_{\alpha}}(z^s_t, t), \epsilon^i_t\Bigr)\\
    &\hspace{.2in}+ (\theta_{\alpha + 1} - \theta_{\alpha})\nabla_{\theta_{\alpha}}\mathcal{L}\Bigl(f_{\theta_{\alpha}}(z^s_t, t), \epsilon^i_t\Bigr) + O(||\theta_{\alpha + 1} - \theta_{\alpha}||^2)\notag
\end{align}
}

Given the small magnitude of the learning rate $\eta$, we disregard the higher-order term $O(||\theta_{\alpha + 1} - \theta_{\alpha}||^2)$, as it scales with $O(||\eta_\alpha||^2)$ and is therefore negligible. Then we have:
{\small
\begin{align}
    \hspace{-.05in}\mathcal{I}_{\theta_{\alpha + 1}, t}(x^s, x^i) &= \mathcal{L}\Bigl(f_{\theta_{\alpha + 1}}(z^s_t, t), \epsilon^i_t\Bigr) - \mathcal{L}\Bigl(f_{\theta_{\alpha}}(z^s_t, t), \epsilon^i_t\Bigr)\\
    &\Rightarrow \eta_{\alpha}\nabla_{\theta_{\alpha}}\mathcal{L}\Bigl(f_{\theta_{\alpha}}(z^i_t, t), \epsilon^i_t\Bigr)\nabla_{\theta_{\alpha}}\mathcal{L}\Bigl(f_{\theta_{\alpha}}(z^s_t, t), \epsilon^i_t\Bigr)\notag
\end{align}
}

To estimate the influence of training data sample $x^i$, we sum over all training iterations that use $x^i$ and over timesteps $t \in \{1, 2, \cdots, T\}$:
{\small
\begin{align}
    &\hspace{-.1in}\mathcal{I}_{\theta^*}(x^s, x^i) = \\
    &\sum_{\theta_a\text{:} x^i}\sum_{t=1}^T\eta_{\alpha}\nabla_{\theta_{\alpha}}\mathcal{L}\Bigl(f_{\theta_{\alpha}}(z^i_t, t), \epsilon^i_t\Bigr)\nabla_{\theta_{\alpha}}\mathcal{L}\Bigl(f_{\theta_{\alpha}}(z^s_t, t), \epsilon^i_t\Bigr)\notag
\end{align}
}where $\theta_a\text{:}x^i$ denotes training iterations that use $x^i$. However, it is impractical to store model parameters and the Gaussian noise for each training iteration. Thus, for a diffusion model with parameters $\theta$, given a test generation $x^s$, we estimate the influence of a training data sample $x^i$ with respect to $x^s$ by:
{\small
\begin{align}
\label{equ:influence_estimation_x} \hspace{-.1in}\mathcal{I}_{\theta}(x^s, x^i) = e\bar{\eta}\sum_{t=1}^T\nabla_{\theta}\mathcal{L}\Bigl(f_{\theta}(z^i_t, t), \epsilon\Bigr)\nabla_{\theta}\mathcal{L}\Bigl(f_{\theta}(z^s_t, t), \epsilon\Bigr)
\end{align}
}where $e$ is the number of epochs, $\bar{\eta}$ is the average learning rate during training, $\epsilon$ corresponds to the Gaussian noise used in the training process. However, storing all Gaussian noise from the training process is impractical. Therefore, for influence estimation we draw Gaussian noise from the same distribution as in training.

Similarly, for a text-to-image model, the influence of a training data sample $X^i=(p^i, x^i)$ with respect to test generation $X^s =(p^s, x^s)$ can be estimated by:
{\small
\begin{align}
\label{equ:influence_estimation_X}
    \hspace{-.1in}\mathcal{I}_{\theta}(X^s, X^i) = e\bar{\eta}\sum_{t=1}^T\nabla_{\theta}\mathcal{L}\Bigl(f_{\theta}(z^i_p, z^i_t, t), \epsilon\Bigr)\nabla_{\theta}\mathcal{L}\Bigl(f_{\theta}(z^s_p, z^s_t, t), \epsilon\Bigr)
\end{align}
}where $p^i$ is the prompt of the training data sample, $p^s$ denotes the prompt of the test generation, and $z^i_p$ and $z^s_p$ are the embeddings of prompts $p^i$ and $p^s$, respectively.

The summation over timesteps in Equation~\ref{equ:influence_estimation_X} should be interpreted as a first-order approximation of the training trajectory, rather than an assumption that timestep contributions are statistically independent. In this formulation, cross-timestep dependencies are implicitly reflected in the final parameter state $\theta$ where all gradients are evaluated. For scalability, we omit second-order cross-timestep terms, consistent with first-order influence estimation methods.

\section{\texttt{DMin}: Scalable Influence Estimation}
For a given generated image $x^s$ and the corresponding prompt $p^s$, the objective of \texttt{DMin} is to estimate an influence score $\mathcal{I}_{\theta}(X^s, X^i)$ for each training pair $X^i = (p^i, x^i)$, where $X^s = (p^s, x^s)$. Based on Equation~\ref{equ:influence_estimation_X}, $\mathcal{I}_{\theta}(X^s, X^i)$ can be expressed as the sum of inner products between the loss gradients of the training sample and the generated image, computed with respect to the same noise $\epsilon$ across timesteps $t\in \{1, 2, \cdots, T\}$. Since the training dataset is fixed and remains unchanged after training, a straightforward approach is to cache or store the gradients of each training sample across timesteps. When estimating the influence for a given query generated image, we only need to compute the gradient for the generated image and perform inner product with the cached gradients of each training sample.

However, as the size of diffusion models and training datasets grows, simply caching the gradients becomes infeasible due to the immense storage requirements. For instance, for a diffusion model with 2B parameters and $1,000$ timesteps, caching the loss gradient of a single training sample would require over $7,450$ GB of storage, making the approach impractical when scaled to large datasets.

In this section, we explain how we reduce the storage requirements for caching such large gradients from gigabytes to kilobytes (Gradient Computation) and how we perform influence estimation for a given generated image on the fly (Influence Estimation), as shown in Figure~\ref{fig:overview}. We use Stable Diffusion on the text-to-image task as an example in this section; similar procedures can be applied to other models.

\subsection{Gradient Computation}
Since the training dataset remains fixed after training, we can cache the loss gradient of each training data sample, as illustrated in Figure~\ref{fig:overview}(a). For a given training pair $X^i = (p^i, x^i)$, and a timestep $t$, the training data is processed through the diffusion model in the same way as during training, and a loss is computed between the model-predicted noise and a Gaussian noise $\epsilon$, where $\epsilon \sim \mathcal{N}(0, I)$. Back-propagation is then performed to obtain the gradient $g^i_t$ for the training data pair $X^i$ at timestep $t$. Once all gradients $\{g^i_1, g^i_2, \cdots, g^i_T\}$ for $X^i$ at all timesteps are obtained, we apply a compression technique to these gradients and cache the compressed versions for influence estimation. Furthermore, for tasks where only the top-$k$ most influential samples are required, we can construct a KNN index on the compressed gradients to enable efficient querying.

\textbf{Forward and Backward Passes.}
In the forward pass, following the same process as training, for a training pair $(p^i, x^i)$ and a timestep $t$, the prompt $p^i$ is passed through the encoder to obtain a prompt embedding, while the image $x^i$ is passed through a VAE to obtain a latent representation $z_0^i$. Gaussian noise is then sampled from $\epsilon \sim \mathcal{N}(0, I)$ and added to the latent representation to obtain a noisy latent representation. The timestep $t$, the noisy latent $z^i_t=z_0^i+\epsilon$, and the embedding $z_p^i$ are then fed into the model for the forward pass.

After the forward pass, a loss is computed between the Gaussian noise $\epsilon$ and the predicted noise $\hat{\epsilon}$. Subsequently, back-propagation is performed to calculate the gradients for each parameter that requires a gradient. It is important to note that for models with adapters as illustrated in Figure~\ref{fig:overview}(b), only the parameters associated with the adapters require gradients. After obtaining the gradients, we concatenate all of them and flatten them into a single vector. For a diffusion model with 2B parameters, this resulting gradient vector will have a length of 2B.

The number of training timesteps is typically 1,000, depending on the model training configuration. For a single training data sample, using a diffusion model with 2B parameters as an example, computing gradients for all $1,000$ timesteps is computationally intensive and costly, requiring over $7,450$ GB of storage. To mitigate this, similar to the inference process in diffusion models, we can sample a subset of timesteps from $t\in \{1, 2, \cdots, T\}$ instead of computing gradients for all timesteps, substantially reducing the computational and storage burden.

\textbf{Gradient Compression.}
However, even storing the gradient vector for a single training data sample at just one timestep requires approximately 7 GB of storage. This becomes impractical for extremely large training datasets containing millions of samples. Therefore, gradient compression techniques are essential to enable caching gradients at this scale efficiently.

As previously discussed, some prior studies employ random projection to compress gradient vectors. However, for a diffusion model with 2B parameters, such compression requires a projection matrix of size $2\text{B} \times v$, where $v$ is the dimension after compression. Even with a modest $v=4096$, this matrix would require over 29 TB of storage. This makes these approaches feasible only for small models or LoRA-tuned models, substantially limiting their scalability.

Inspired by prior work on vector compression~\cite{DBLP:conf/kdd/0001L23, DBLP:conf/acl/LinLX024}, we compress the gradient vector through four steps: (1) padding, (2) permutation, (3) random projection, and (4) group addition. In the gradient compression process, we first pad the gradient vector to the smallest length $L_\text{pad}$ that can be evenly divided by $v$. Padding can be achieved by appending 0s to the original gradient vector until the desired length is reached. Next, we permute the gradient vector using a random permutation to disrupt any inherent structure in the vector representation. We then perform an element-wise multiplication of the permuted gradient vector with a random projection vector. The random projection vector is of the same length as the gradient vector and consists of elements randomly set to either -1 or 1 with equal probability. This step projects the gradient onto a randomized basis, reducing redundancy while preserving essential information. Finally, we divide the $L_\text{pad}$ elements of the gradient vector into $\frac{L_\text{pad}}{v}$ groups, summing up the elements within each group to produce the compressed vector of dimension $v$.

With this compression, we only need to store two components: a permutation vector that records the indices of the permutation (4 bytes per element) and a binary projection vector (1 bit per element). As a result, the storage requirement is significantly reduced, occupying just 7.45 GB for the gradients plus an additional 238 MB for the projection vector. This reduction makes it feasible to store and cache the gradients for influence estimation at scale.

\textbf{Normalization.}
Some prior studies have highlighted the inherent instability of gradients in deep learning~\cite{DBLP:conf/acl/LinLX024, DBLP:conf/iclr/BasuPF21, DBLP:journals/nn/EpifanoRMR23, DBLP:conf/aaai/GhorbaniAZ19}, particularly in extremely large models. This instability arises from the potential for unusually large weights and gradients in the model. In our experiments, we encountered this issue: the magnitude of some gradient values is found to be extremely large. Such large gradient values can dominate the inner product, leading to incorrect results. To address this, we apply L2 normalization to the gradient vector before compression, which effectively mitigates the impact of unusually large gradient magnitudes. Consequently, Equation~\ref{equ:influence_estimation_X} can be reformulated as:
{
\small
\begin{align}
&\hspace{-.05in}\mathcal{I}_{\theta}(X^s, X^i) = \\
&\hspace{.1in}e\bar{\eta}\sum_{t=1}^T 
    \frac{\nabla_{\theta}\mathcal{L}\Bigl(f_{\theta}(z^i_p, z^i_t, t), \epsilon\Bigr)}{\bigl\|\nabla_{\theta}\mathcal{L}\Bigl(f_{\theta}(z^i_p, z^i_t, t), \epsilon\Bigr)\bigr\|_2} \cdot \frac{\nabla_{\theta}\mathcal{L}\Bigl(f_{\theta}(z^s_p, z^s_t, t), \epsilon\Bigr)}{\bigl\|\nabla_{\theta}\mathcal{L}\Bigl(f_{\theta}(z^s_p, z^s_t, t), \epsilon\Bigr)\bigr\|_2} \notag
\end{align}
}

\textbf{Index Construction for KNN.}
To further enhance the scalability of \texttt{DMin}, we introduce KNN search for tasks requiring only the top-$k$ most influential samples. After gradient compression, as shown in Figure~\ref{fig:overview}(d), we concatenate all the compressed gradients across timesteps to construct a KNN index, enabling efficient querying during influence estimation. This approach is well-suited for extremely large datasets, allowing for the retrieval of the top-$k$ most influential samples on the fly.

\subsection{Influence Estimation}
After caching the compressed gradients, for a given generated image and its corresponding prompt, we compute and compress the gradient in the same way as for the training data samples to obtain the compressed gradient for the given sample. For exact influence estimation, we calculate the inner product between the compressed gradient of the given sample and the cached compressed gradients of each training sample across timesteps to obtain the influence scores. For KNN retrieval, we concatenate the compressed gradients across timesteps to query the KNN index and identify the top-$k$ most relevant training samples efficiently.

\begin{table}[t]
\centering
\resizebox{.35\textwidth}{!}{%
\begin{tabular}{lccc}
\toprule
\midrule
Subset & \# Train & \% of Training Data & \# Test\\\midrule
Flowers & 162 & 1.74\% & 34 \\
Lego Sets & 40 & 0.43\% & 21 \\
Magic Cards & 1541 & 16.59\% & 375 \\
\midrule
\bottomrule
\end{tabular}%
}
\vspace{-.1in}
\caption{Sub-datasets used in experimental evaluation. (Full dataset is listed in Appendix~\ref{apd:datasets}.)}
\label{tbl:3_subsets}
\vspace{-.1in}
\end{table}

\section{Experiments}

In this section, we present our experiments conducted on various models and settings to validate the effectiveness and efficiency of the proposed \texttt{DMin}.

\textbf{Datasets.}
For the conditional diffusion model, we combine six datasets from Hugging Face and randomly select 80\% of the data samples as the training dataset, resulting in 9,288 pairs of images and prompts. Due to page limitations, we list three datasets used for evaluation in Table~\ref{tbl:3_subsets}: (1) Flowers, which includes 162 training pairs of flower images and corresponding descriptive prompts in our experiments, accounting for only 1.74\% of the training dataset. (2) Lego Sets: This subset consists of 40 training pairs, where each image represents a Lego box accompanied by a description of the box, accounting for only 0.43\% of the training dataset. (3) Magic Cards, which contains magic card images from Scryfall with captions generated by Fuyu-8B~\cite{fuyu-8b} and BLIP~\cite{DBLP:conf/icml/0001LXH22}. For unconditional diffusion models, we mainly focus on MNIST and CIFAR-10. We include a detailed explanation of datasets\footnotemark[5]\footnotemark[6]\footnotemark[7] in Appendix~\ref{apd:datasets}.


\textbf{Models.}
For conditional text-to-image diffusion models, we use three different models: (1) SD 1.4 with LoRA, (2) SD 3 Medium with LoRA and (3) SD 3 Medium (Full parameters). For unconditional diffusion models, we conduct experiments on two Denoising Diffusion Probabilistic Models (DDPM) trained on MNIST and CIFAR-10. The detailed settings of models are included in Appendix~\ref{apd:models}. We fine-tune models on the combined training dataset mentioned above and evaluate them on the testing dataset. During gradient collection, we collect only the gradients of the parameters in the LoRA components for the LoRA-tuned model, whereas for the fully fine-tuned model, we collect the gradients of all parameters (Figure~\ref{fig:overview}(b)).


\begin{table*}[t]
\vspace{-.15in}
\centering
\caption{Average detection rates of top-k most influential training data samples. $\text{Detection rate} = \frac{\text{\# Samples from Same Subset among Top-k Training Samples}}{k}$ where $k=\{5, 10, 50, 100\}$, indicating the average proportion of samples from the same subset appearing in the top-k influential samples. ``Ours (w/o Comp.)'' indicates that the gradient vectors are not compressed, while ``w/o Norm.'' signifies that the gradient vectors are not normalized. ``Excatly'' denotes exact inner product computation. The results for LiSSA, DataInf and D-TRAK on SD3 Medium (Full) are omitted due to hundreds of TB of cache. Moreover, at this scale, it is impractical for LiSSA and DataInf to approximate the Hessian inversion and for D-TRAK to compute a large random projection matrix.}
\vspace{-.1in}
\resizebox{.86\textwidth}{!}{%
\begin{tabular}{cl|cccc|cccc|cccc}
\toprule
\midrule
\multirow{2}{*}{Model}                                                         & \multirow{2}{*}{Method}                       & \multicolumn{4}{c|}{Flowers}                                           & \multicolumn{4}{c|}{Lego Sets}                                         & \multicolumn{4}{c}{Magic Cards}                                       \\
                                                                               &                                               & Top 5           & Top 10          & Top 50          & Top 100         & Top 5           & Top 10          & Top 50          & Top 100         & Top 5           & Top 10          & Top 50          & Top 100         \\\midrule
\multirow{13}{*}{\begin{tabular}[c]{@{}c@{}}SD 1.4\\ (LoRA)\end{tabular}}      & Random Selection                              & 0.0000          & 0.0000          & 0.0200          & 0.0100          & 0.0000          & 0.0000          & 0.0000          & 0.0000          & 0.2000          & 0.2000          & 0.0800          & 0.1300          \\
                                                                               & SSIM                                          & 0.2000          & 0.1000          & 0.0220          & 0.0130          & 0.0400          & 0.0400          & 0.0340          & 0.0240          & 0.2800          & 0.3500          & 0.4480          & 0.4290          \\
                                                                               & CLIP Similarity                               & 0.0000          & 0.0000          & 0.0000          & 0.0000          & 0.0000          & 0.0000          & 0.0000          & 0.0000          & 0.4444          & 0.4005          & 0.3565          & 0.3830          \\
                                                                               & LiSSA                                         & 0.5143          & 0.4571          & 0.3486          & 0.2929          & 0.0000          & 0.0000          & 0.0040          & 0.0080          & 0.9667          & 0.9500          & 0.9600          & 0.9483          \\
                                                                               & DataInf (Identity)                            & 0.4125          & 0.4062          & 0.3188          & 0.2687          & 0.0000          & 0.0000          & 0.0067          & 0.0100          & 0.9667          & 0.9500          & 0.9600          & 0.9483          \\
                                                                               & DataInf (Hessian Inversion)                   & 0.4125          & 0.4062          & 0.3188          & 0.2687          & 0.0000          & 0.0000          & 0.0067          & 0.0100          & 0.9667          & 0.9500          & 0.9600          & 0.9483          \\
                                                                               & Ours (w/o Comp. \& Norm.)                     & 0.1333          & 0.1154          & 0.1138          & 0.1028          & 0.0000          & 0.0000          & 0.0047          & 0.0065          & 0.9637          & 0.9585          & 0.9402          & 0.9280          \\
                                                                               & Ours (w/o Comp.)                              & \textbf{0.8872} & \textbf{0.8359} & \textbf{0.5836} & \textbf{0.3969} & \textbf{0.5647} & \textbf{0.4412} & 0.1435          & \textbf{0.0894} & \textbf{0.9778} & 0.9778          & 0.9911          & 0.9933          \\
                                                                               & Ours ($v=2^{12}$, Exactly)                    & 0.8667          & 0.8154          & 0.5713          & 0.3836          & 0.5176          & 0.3882          & 0.1435          & 0.0865          & \textbf{0.9778} & \textbf{0.9889} & \textbf{0.9933} & \textbf{0.9944} \\
                                                                               & Ours ($v=2^{16}$, Exactly)                    & 0.8615          & 0.8231          & 0.5718          & 0.3813          & 0.5529          & 0.4353          & \textbf{0.1447} & \textbf{0.0894} & \textbf{0.9778} & 0.9778          & 0.9911          & 0.9933          \\
                                                                               & Ours ($v=2^{20}$, Exactly)                    & 0.8667          & 0.8154          & 0.5713          & 0.3836          & \textbf{0.5647} & \textbf{0.4412} & 0.1435          & \textbf{0.0894} & \textbf{0.9778} & 0.9778          & 0.9911          & 0.9933          \\
                                                                               & Ours ($v=2^{12}$, KNN)                        & 0.8615          & 0.8128          & 0.5405          & 0.3585          & 0.5059          & 0.3647          & 0.1365          & 0.0824          & \textbf{0.9778} & \textbf{0.9889} & \textbf{0.9933} & \textbf{0.9944} \\
                                                                               & Ours ($v=2^{16}$, KNN)                        & 0.8615          & 0.8231          & 0.5723          & 0.3808          & 0.5412          & 0.4176          & 0.1388          & 0.0847          & \textbf{0.9778} & \textbf{0.9889} & 0.9889          & \textbf{0.9944} \\\midrule
\multirow{13}{*}{\begin{tabular}[c]{@{}c@{}}SD 3 Medium\\ (LoRA)\end{tabular}} & Random Selection                              & 0.0000          & 0.0000          & 0.0200          & 0.0100          & 0.0000          & 0.0000          & 0.0000          & 0.0000          & 0.2000          & 0.2000          & 0.0800          & 0.1300          \\
                                                                               & SSIM                                          & 0.1800          & 0.0900          & 0.0200          & 0.0160          & 0.0000          & 0.0000          & 0.0160          & 0.0190          & 0.0000          & 0.0067          & 0.0180          & 0.0347          \\
                                                                               & CLIP Similarity                               & 0.0000          & 0.0000          & 0.0000          & 0.0000          & 0.0000          & 0.0000          & 0.0000          & 0.0000          & 0.0352          & 0.0363          & 0.0421          & 0.0438          \\
                                                                               & LiSSA                                         & 0.8889          & 0.8889          & 0.8622          & 0.8222          & 0.1111          & 0.1111          & 0.1244          & 0.1044          & 0.9091          & 0.9091          & 0.9091          & 0.9082          \\
                                                                               & DataInf (Identity)                            & 0.8556          & 0.8556          & 0.7878          & 0.6683          & 0.1647          & 0.1176          & 0.0576          & 0.0424          & 0.8833          & 0.8917          & 0.8900          & 0.8883          \\
                                                                               & DataInf (Hessian Inversion)                   & 0.8556          & 0.8556          & 0.7878          & 0.6683          & 0.1647          & 0.1176          & 0.0576          & 0.0424          & 0.8833          & 0.8917          & 0.8900          & 0.8883          \\
                                                                               & Ours (w/o Comp. \& Norm.)                     & 0.8974          & 0.8769          & 0.8010          & 0.6738          & 0.2588          & 0.1765          & 0.1024          & 0.0765          & 0.7935          & 0.7951          & 0.7965          & 0.7986          \\
                                                                               & Ours (w/o Comp.)                              & \textbf{0.9128} & 0.8974          & 0.8390          & 0.7605          & 0.6118          & 0.5059          & 0.2318          & 0.1488          & \textbf{1.0000} & \textbf{1.0000} & \textbf{1.0000} & 0.9700          \\
                                                                               & Ours ($v=2^{12}$, Exactly)                    & 0.8974          & 0.8846          & 0.8318          & 0.7608          & 0.6000          & 0.5235          & 0.2306          & 0.1529          & 0.9837          & 0.9835          & 0.9751          & 0.9703          \\
                                                                               & Ours ($v=2^{16}$, Exactly)                    & 0.9077          & 0.8872          & 0.8405          & 0.7659          & 0.5765          & 0.5118          & 0.2224          & 0.1482          & 0.9848          & 0.9840          & 0.9761          & 0.9718          \\
                                                                               & Ours ($v=2^{20}$, Exactly)                    & 0.9077          & 0.8872          & 0.8385          & 0.7651          & 0.6000          & 0.5235          & 0.2294          & 0.1506          & 0.9848          & 0.9840          & 0.9762          & 0.9720          \\
                                                                               & Ours ($v=2^{12}$, KNN)                        & 0.9026          & 0.8949          & 0.8415          & 0.7641          & \textbf{0.7294} & \textbf{0.6529} & \textbf{0.3094} & \textbf{0.1924} & 0.9854          & 0.9851          & 0.9771          & 0.9717          \\
                                                                               & Ours ($v=2^{16}$, KNN)                        & \textbf{0.9128} & \textbf{0.9051} & \textbf{0.8472} & \textbf{0.7721} & 0.7059          & 0.6353          & 0.3035          & 0.1871          & 0.9864          & 0.9862          & 0.9785          & \textbf{0.9736} \\\midrule
\multirow{8}{*}{\begin{tabular}[c]{@{}c@{}}SD 3 Medium\\ (Full)\end{tabular}}  & Random Selection                              & 0.0000          & 0.0000          & 0.0200          & 0.0100          & 0.0000          & 0.0000          & 0.0000          & 0.0000          & 0.2000          & 0.2000          & 0.0800          & 0.1300          \\
                                                                               & SSIM                                          & 0.1800          & 0.0967          & 0.0200          & 0.0117          & 0.0235          & 0.0176          & 0.0282          & 0.0206          & 0.0000          & 0.0000          & 0.0020          & 0.0160          \\
                                                                               & CLIP Similarity                               & 0.0000          & 0.0000          & 0.0000          & 0.0000          & 0.0000          & 0.0000          & 0.0000          & 0.0000          & 0.2938          & 0.3412          & 0.4583          & 0.4982          \\
                                                                               & Ours ($v=2^{12}$, Exactly) & 0.9487          & 0.9000          & 0.5385          & 0.3567          & 0.5529          & 0.4412          & 0.1906          & 0.1165          & 0.9882          & 0.9882          & 0.9420          & 0.9063          \\
                                                                               & Ours ($v=2^{16}$, Exactly) & 0.9590          & 0.9308          & 0.5564          & 0.3690          & 0.5765          & 0.4765          & 0.2047          & 0.1282          & \textbf{0.9961} & 0.9902          & 0.9514          & 0.9220          \\
                                                                               & Ours ($v=2^{20}$, Exactly) & \textbf{0.9641} & \textbf{0.9333} & 0.5590          & \textbf{0.3708} & 0.5647          & 0.4765          & 0.2071          & 0.1306          & 0.9922          & 0.9922          & 0.9498          & 0.9202          \\
                                                                               & Ours ($v=2^{12}$, KNN)     & 0.9282          & 0.8641          & 0.5354          & 0.3518          & 0.6125          & 0.5062          & 0.2025          & 0.1288          & 0.9880          & 0.9820          & 0.9472          & 0.9046          \\
                                                                               & Ours ($v=2^{16}$, KNN)     & 0.9622          & 0.9108          & \textbf{0.5622} & 0.3695          & \textbf{0.6250} & \textbf{0.5437} & \textbf{0.2213} & \textbf{0.1419} & 0.9960          & \textbf{0.9960} & \textbf{0.9640} & \textbf{0.9308}
\\
 \midrule
 \bottomrule
\end{tabular}%
}
\label{tbl:overall_performance}
\vspace{-.15in}
\end{table*}

\textbf{Baselines.}
We compare the proposed \texttt{DMin} against the following baselines:
(1) Random Selection: Assigns an influence score to each training sample randomly.
(2) SSIM~\cite{DBLP:journals/tip/BrunetVW12}: Structural Similarity Index Measure (SSIM) between the training image and the generated image.
(3) CLIP Similarity~\cite{DBLP:conf/icml/RadfordKHRGASAM21}: Cosine similarity of embeddings computed by CLIP between the training image and the generated image.
(4) LiSSA~\cite{DBLP:journals/jmlr/AgarwalBH17}: A second-order influence estimation method that uses an iterative approach to compute the inverse Hessian-vector product.
(5) DataInf~\cite{DBLP:conf/iclr/KwonWW024}: An influence estimation method based on a closed-form expression for computational efficiency. We also evaluate a variant of DataInf where the Hessian Inversion matrix is replaced with an identity matrix.
(6) D-TRAK~\cite{DBLP:conf/emnlp/OguejiAOGHK22}: A first-order influence estimation method extended from TRAK~\cite{DBLP:conf/icml/ParkGILM23}.
(7) Journey-TRAK~\cite{DBLP:journals/corr/abs-2312-06205}: An estimation method focusing on the sampling path in diffusion models.
For the proposed \texttt{DMin}, we evaluate \texttt{DMin} under different scenarios, including exact estimation of influence scores for each training sample and KNN-based approximate searches for the top-$k$ most influential samples. Additionally, we experiment with varying compression levels: no compression, and $v=\{2^{12}, 2^{16}, 2^{20}\}$. Details of the baselines are reported in Appendix~\ref{apd:baselines}.

\textbf{KNN.} We use the hierarchical navigable small world (HNSW) algorithm~\cite{DBLP:journals/pami/MalkovY20} for KNN in our experiments, and we provide the results of an ablation study in Appendix~\ref{apd:ablation}.

\subsection{Performance on Conditional Diffusion Models}
The goal of this experiment is to confirm the effectiveness of different methods in identifying influential training samples within the training dataset. 

\textbf{Visualization.} Figure~\ref{fig:top_figure} illustrates several examples, showing the generated image and its corresponding prompt in the first column, followed by the training samples ranked from highest to lowest influence, arranged from left to right. These examples demonstrate that the proposed \texttt{DMin} method successfully retrieves training image samples with content similar to the generated image and prompt. Additional visualizations are provided in Appendix~\ref{apd:vis_conditional}.

\begin{figure*}[t]
\vspace{-.1in}
    \centering
    \includegraphics[width=0.95\linewidth]{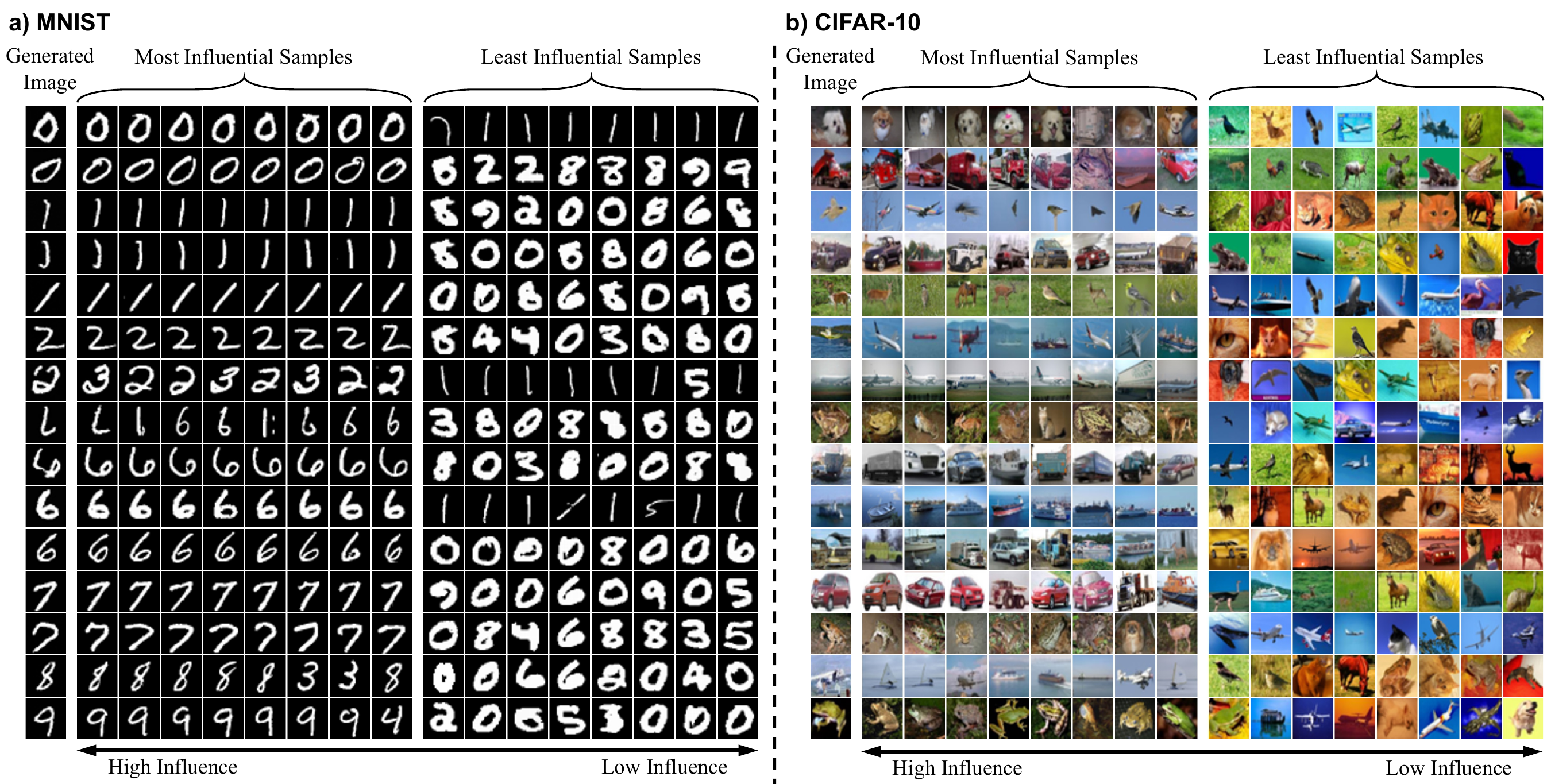}
    \vspace{-.12in}
    \caption{Examples of generated images alongside the most and least influential samples (from left to right) as estimated by \texttt{DMin} for unconditional DDPM models on the MNIST and CIFAR-10 datasets.}
    \label{fig:examples_unconditional}  
    \vspace{-.15in}
\end{figure*}

\textbf{Qualitative Analysis.} Unlike prior studies focusing on small diffusion models, the diffusion models used in our experiments are substantially larger, making it impractical to retrain them for leave-one-out evaluation. Consequently, we assess the detection rate in our experiments, as shown in Table~\ref{tbl:overall_performance}, which reflects the average proportion of similar content from the training dataset retrieved by the top-$k$ most influential samples.

\textbf{Datasets.} As mentioned earlier, our training dataset is a combination of six datasets. As shown in Table~\ref{tbl:3_subsets}, we report evaluations on three subsets: Flowers, Lego Sets, and Magic Cards, as these subsets are more distinct from the others. For example, given a prompt asking the model to generate a magic card, the generated image should be more closely related to the Magic Cards subset rather than the Flowers or Lego Sets subsets, as the knowledge required to generate magic cards primarily originates from the Magic Cards subset. Similarly, the knowledge for generating images containing Lego comes predominantly from the Lego Sets subset. Therefore, for a prompt belonging to one of the test subsets -- Flowers, Lego Sets, or Magic Cards -- the most influential training samples are highly likely to originate from the same subset. This implies that a greater number of training samples from the corresponding subset should be identified among the top-$k$ most influential samples.

We begin by generating images using the prompts from the test set of each subset -- Flowers, Lego Sets, and Magic Cards. 
For each test prompt and its generated image, we estimate the influence score for every training data sample and select the top-$k$ most influential training samples with the highest influence score.
We then calculate the detection rate as $\text{Detection Rate} = \frac{\text{\# Samples from Same Subset among top-$k$ Training Samples}}{k}$.

\textbf{Results.} We report the average detection rate for each test set of subsets in Table~\ref{tbl:overall_performance}.
Compared to the baselines, our proposed \texttt{DMin} achieves the best performance across all subsets. The detection rates for top-50 and top-100 on Lego Sets are lower because the Lego Sets training dataset contains only 40 samples (0.43\% of the total). Across all subsets and different values of $k$, $v=2^{16}$ achieves the best performance in most cases, whether using KNN or exact inner product computation. Additionally, compared to our method without compression, removing normalization substantially decreases performance, confirming that normalization mitigates the instability of gradients in extremely deep models. Interestingly, KNN search often outperforms exact inner product computation in our experiments across all models and subsets. This improvement is likely due to KNN's ability to approximate the search process, capturing a broader and more representative subset of neighbors.

\begin{table*}[t]
\centering
\vspace{-.15in}
\caption{Storage requirements for caching per-sample and dataset gradients (9,288 samples), comparing compressed and uncompressed methods across models. The table shows storage and compression ratios of our method across levels, with LiSSA and DataInf storing gradients uncompressed.}
\vspace{-.1in}
\resizebox{.82\linewidth}{!}{%
\begin{tabular}{l|ccc|ccc|ccc}
\toprule
\midrule
\multirow{2}{*}{Mehod} & \multicolumn{3}{c|}{SD 1.4 (LoRA, 10 Timesteps)} & \multicolumn{3}{c|}{SD 3 Medium (LoRA, 10 Timesteps)} & \multicolumn{3}{c}{SD 3 Medium (Full, 5 Timesteps)} \\
 & \begin{tabular}[c]{@{}c@{}}Size\\ (Per Sample)\end{tabular} & \begin{tabular}[c]{@{}c@{}}Size\\ (Training Dataset)\end{tabular} & \begin{tabular}[c]{@{}c@{}}Compression\\ Ratio\end{tabular} & \begin{tabular}[c]{@{}c@{}}Size\\ (Per Sample)\end{tabular} & \begin{tabular}[c]{@{}c@{}}Size\\ (Training Dataset)\end{tabular} & \begin{tabular}[c]{@{}c@{}}Compression\\ Ratio\end{tabular} & \begin{tabular}[c]{@{}c@{}}Size\\ (Per Sample)\end{tabular} & \begin{tabular}[c]{@{}c@{}}Size\\ (Training Dataset)\end{tabular} & \begin{tabular}[c]{@{}c@{}}Compression\\ Ratio\end{tabular} \\\midrule
Gradient w/o Comp. & 30.41 MB & 275.82 GB & 100\% & 45 MB & 408.16 GB & 100\% & 37.42 GB & 339.39 TB & 100\% \\
Ours ($v=2^{12}$) & 160 KB & 1.45 GB & 0.53\% & 160 KB & 1.45 GB & 0.36\% & 80 KB & 726 MB & 0.00017\% \\
Ours ($v=2^{16}$) & 2.5 MB & 22.68 GB & 8.22\% & 2.5 MB & 22.68 GB & 5.56\% & 1.25 MB & 11.34 GB & 0.0028\% \\
Ours ($v=2^{20}$) & - & - & - & - & - & - & 20 GB & 181.41 GB & 0.044\% \\
\midrule
\bottomrule
\end{tabular}%
}
\vspace{-.1in}
\label{tbl:memory}

\vspace{-.1in}
\end{table*}
\begin{table}[t]
\centering
\caption{Time cost comparison and speedup relative to our method without compression. Time cost refers to the time to estimate influence scores for all training samples per test sample (in seconds).}
\vspace{-.1in}
\resizebox{1\linewidth}{!}{%
\begin{tabular}{l|cc|cc|c}
\toprule
\midrule
\multirow{3}{*}{Mehod} & \multicolumn{2}{c|}{SD 1.4 (LoRA)} & \multicolumn{2}{c|}{SD 3 Medium (LoRA)} & SD 3 Medium (Ful) \\
 & \begin{tabular}[c]{@{}c@{}}Time\\ {\small (seconds/test sample)}\end{tabular} & \begin{tabular}[c]{@{}c@{}}Speedup\\ {\small (vs. w/o Comp.)}\end{tabular} & \begin{tabular}[c]{@{}c@{}}Time\\ {\small (seconds/test sample)}\end{tabular} & \begin{tabular}[c]{@{}c@{}}Speedup\\ {\small (vs. w/o Comp.)}\end{tabular} & \begin{tabular}[c]{@{}c@{}}Time\\ {\small (seconds/test sample)}\end{tabular} \\\midrule
LiSSA & 2,939.283 & 0.02x & 2,136.701 & 0.19x & - \\
DataInf {\small(Identity)} & 206.385 & 0.34x & 201.923 & 2.02x & - \\
DataInf {\small(Hessian Inversion)} & 1,187.841 & 0.06x & 932.762 & 0.44x & - \\
D-TRAK & 345.223 & 0.20x & 833.850 & 0.49x & - \\\midrule
Ours {\small(w/o Comp.)} & 70.590 & 1x & 407.511 & 1x & - \\
Ours {\small($v=2^{12}$, Exact)} & 8.193 & 8.62x & 14.238 & 28.62x & 9.866 \\
Ours {\small($v=2^{16}$, Exact)} & 41.026 & 1.72x & 135.462 & 3.01x & 18.900 \\
Ours {\small($v=2^{20}$, Exact)} & 99.307 & 0.71x & 623.610 & 0.65x & 100.880 \\\midrule
Ours {\small($v=2^{12}$, KNN, Top-5)} & \textbf{0.004} & \textbf{18,100.51x} & \textbf{0.004} & \textbf{101,877.75x} & \textbf{0.009} \\
Ours {\small($v=2^{12}$, KNN, Top-50)} & 0.018 & 3,921.78x & 0.010 & 40,751.10x & 0.014 \\
Ours {\small($v=2^{12}$, KNN, Top-100)} & 0.033 & 2,139.15x & 0.019 & 21,447.95x & 0.131 \\\midrule
Ours {\small($v=2^{16}$, KNN, Top-5)} & 0.073 & 967.01x & 0.065 & 6,269.40x & 0.097 \\
Ours {\small($v=2^{16}$, KNN, Top-50)} & 0.393 & 179.62x & 0.227 & 1,792.04x & 0.485 \\
Ours {\small($v=2^{16}$, KNN, Top-100)} & 0.736 & 95.91x & 0.406 & 1,003.72x & 0.784 \\
\midrule
\bottomrule
\end{tabular}%
}
\label{tbl:time}
\vspace{-.15in}
\end{table}

\subsection{Time and Memory Cost}\label{sec:time_memory_cost}
The computational cost of both time and memory is critical for evaluating the scalability of influence estimation methods, especially when applied to large diffusion models. 

\textbf{Time.} Table~\ref{tbl:time} reports the time required to estimate the influence score for every training sample in the training dataset for a single test sample. The gradient computation and caching times for the two LoRA-tuned models are similar across methods: (1) SD 1.4 (LoRA): around 8 GPU hours, (2) SD 3 Medium (LoRA): around 24 GPU hours, and (3) SD 3 Medium (full): 330 GPU hours. Additionally, the index construction process only takes a few minutes. Our proposed methods demonstrate substantial efficiency improvements, particularly with KNN search. For instance, on the smallest subset—Lego Sets, which contains only 21 test samples—estimating the influence score for the entire training dataset takes 17 hours with LiSSA, 7 hours with DataInf (Hessian Inversion), and 2 hours with D-TRAK. In contrast, our method with $v=2^{12}$ and $k=5$ requires only 0.084 seconds, and even for $k=100$, it takes only 0.69 seconds while achieving the best performance.

\textbf{Memory.} Table~\ref{tbl:memory} compares the storage requirements for caching per-sample gradients and the entire training dataset (9,288 samples) across different models, with and without compression. Without compression, gradient storage is substantially large, reaching 339.39 TB for SD 3 Medium (Full). In contrast, our method achieves drastic reductions in storage size with various compression levels. For example, using $v=2^{12}$, the storage for SD 3 Medium (Full) is reduced to just 726 MB, achieving a compression ratio of 0.00017\%, demonstrating the scalability and efficiency of our approach for handling large-scale models.

\subsection{Unconditional Diffusion Models}
We evaluate the performance of the proposed \texttt{DMin} on unconditional diffusion models using DDPM on the MNIST and CIFAR-10 datasets. Figure~\ref{fig:examples_unconditional} illustrates examples of generated images and the corresponding most and least influential training samples as identified by our method. On MNIST (Figure~\ref{fig:examples_unconditional}(a)), the most influential samples for each generated digit closely resemble the generated image, validating the effectiveness of our approach. Similarly, for CIFAR-10 (Figure~\ref{fig:examples_unconditional}(b)), our method retrieves relevant training samples with similar content. These results highlight the scalability and reliability of our method for detecting influential samples in unconditional diffusion models.

Table~\ref{tbl:mnist} reports the detection rate compared to baseline methods Journey-TRAK and D-TRAK. Our method consistently outperforms both baselines across all metrics, achieving substantially higher detection rates. For instance, with $v=2^{16}$, our method achieves a detection rate of 0.8006 for Top-5 on MNIST, while Journey-TRAK and D-TRAK achieve only 0.2560 and 0.1264, respectively.

\begin{table}[t]
\caption{Detection Rate compared with Journey-TRAK and D-TRAK for the unconditional diffusion model (DDPM) on MNIST.}
\centering
\vspace{-.1in}
\resizebox{0.72\linewidth}{!}{%
\begin{tabular}{lcccc}
\toprule
\midrule
Method & Top 5 & Top 10 & Top 50 & Top 100 \\\midrule
Journey-TRAK & 0.2560 & 0.2190 & 0.1732 & 0.1513 \\
D-TRAK & 0.1264 & 0.1410 & 0.1382 & 0.1272 \\
Ours ($v=2^{12}$, Exact) & 0.4376 & 0.4315 & 0.4094 & 0.4027 \\
Ours ($v=2^{16}$, Exact) & $0.8006$ & $0.7901$ & $0.7408$ & $0.7098$ \\
\midrule
\bottomrule
\end{tabular}%
}
\vspace{-.1in}
\label{tbl:mnist}
\end{table}

\section{Related Work}
Influence estimation has been a critical area of research in understanding the impact of individual training samples on machine learning models~\cite{DBLP:conf/aaai/SchioppaZVS22, DBLP:conf/icml/ParkGILM23, DBLP:journals/corr/abs-2405-17490, DBLP:conf/iclr/ChhabraLM024}. Early work by~\citet{DBLP:conf/icml/KohL17, DBLP:journals/jmlr/AgarwalBH17} proposed second-order Hessian-based methods to approximate the effect of a training sample. However, approximating a Hessian inversion becomes computationally prohibitive for large-scale datasets and modern models containing billions of parameters. To address this issue, some studies proposed first-order approaches for influence estimation~\cite{DBLP:conf/nips/PruthiLKS20, DBLP:conf/icml/ParkGILM23}. However, even with first-order methods, scaling to large datasets still encounters storage challenges. For example, storing the gradient of a 2B diffusion model for 10,000 data samples across 10 timesteps requires over 700 TB of storage.

To reduce the storage and computational demands, some studies leverage dimension reduction techniques~\cite{DBLP:conf/icml/ParkGILM23, DBLP:conf/emnlp/OguejiAOGHK22, DBLP:journals/corr/abs-2312-06205,DBLP:journals/ml/HammoudehL24}, such as random projection. However, while random projection can substantially reduce the dimension of the gradient vector, the projection matrix itself becomes a scalability bottleneck in large models. For instance, in a model with 2B parameters, a projection matrix mapping gradients to a compressed dimension of 32,768 would require over 500 GB of storage. These constraints highlight the need for more efficient and scalable approaches.

\section{Conclusion}
In this paper, we introduce \texttt{DMin}, a scalable framework for estimating the influence of training data samples on images generated by diffusion models. The proposed \texttt{DMin} scales effectively to diffusion models with billions of parameters by substantially reducing storage requirements from hundreds of TBs to MBs or KBs for SD 3 Medium with full parameters. Additionally, \texttt{DMin} can retrieve the top-$k$ most influential training samples in under one second using KNN, demonstrating the scalability of the proposed \texttt{DMin}. Our empirical results further confirm \texttt{DMin}'s effectiveness and efficiency.

\section*{Acknowledgments}

This work was supported by the National Science Foundation under Award Nos. 2543795, 2247619, and 2413046. We acknowledge Research Computing~\cite{RIT_RC} at RIT for providing computing resources. We also acknowledge the Tufts University High Performance Compute Cluster~\cite{tufts_hpc}, which was utilized for the research reported in this paper.

{
    \small
    \bibliographystyle{ieeenat_fullname}
    \bibliography{main}
}


\clearpage
\setcounter{page}{1}
\maketitlesupplementary

\section{Experimental Settings.}
In this section, we report the detailed experimental settings and environments.

\textbf{Implementation Details.}
We provide an open-source PyTorch implementation with multiprocessing support. We leverage Hugging Face, Accelerate, Transformers, Diffusers and PEFT in our implementation.

\textbf{Experimental Environments.}
Our experiments are conducted on three types of servers: (1) Servers running Red Hat Enterprise Linux 7.8, equipped with Intel(R) Xeon(R) Platinum 8358 processors (2.60GHz) with 32 cores, 64 threads, 4 A100 80G GPUs, and 1TB of memory. (2) Servers running Red Hat Enterprise Linux 7.8, containing Intel(R) Xeon(R) Gold 6226R CPUs @ 2.90GHz with 16 cores, 32 threads, 2 A100 40G GPUs, and 754GB of memory. (3) A server running Ubuntu 20.04.6 LTS, featuring 2 H100 GPUs, dual Intel(R) Xeon(R) Gold 6438N processors (3.60GHz) with 32 cores, 64 threads, and 1.48TB of memory. To ensure a fair comparison, all experiments measuring time cost and memory consumption are conducted on server type (1), while other experiments are distributed across the different server types.

\subsection{Models}\label{apd:models}
This study evaluates the performance of the following models:
(1) SD 1.4 with LoRA: This model integrates Stable Diffusion 1.4 (SD 1.4) with Low-Rank Adaptation (LoRA), a technique that fine-tunes large models efficiently by adapting specific layers to the target task while maintaining most of the original model's structure.
(2) SD 3 Medium with LoRA: Utilizing the Stable Diffusion 3 Medium (SD 3 Medium) base model, this configuration applies LoRA for task-specific adaptation. The medium-sized architecture of SD 3 balances computational efficiency with high-quality generation performance.
(3) SD 3 Medium: A standalone version of Stable Diffusion 3 Medium, serving as a baseline for comparison against the LoRA-enhanced models. This version operates without any additional fine-tuning, showcasing the model's capabilities in its default state. Additionally, we include the hyperparameter settings in Table~\ref{tbl:hyperparameters}.

\subsection{Datasets}\label{apd:datasets}

In this section, we introduce the datasets used in our experiments on conditional diffusion models and unconditional diffusion models.

\textbf{Dataset Combination.} For conditional diffusion models, we combine six datasets from Hugging Face: (1) magic-card-captions by clint-greene, (2) midjourney-detailed-prompts by MohamedRashad, (3) diffusiondb-2m-first-5k-canny by HighCWu, (4) lego-sets-latest by merve, (5) pokemon-blip-captions-en-ja by svjack, and (6) gesang-flowers by Albe-njupt. Additionally, we introduced noise to 5\% of the data, selected randomly, and appended it to the dataset to enhance robustness. Finally, we split the data, allocating 80\% (9,288 samples) for training and the remaining 20\% for testing. For unconditional diffusion models, we use two classic datasets: (1) MNIST and (2) CIFAR-10.

\begin{figure*}
    \centering
    \includegraphics[width=\textwidth]{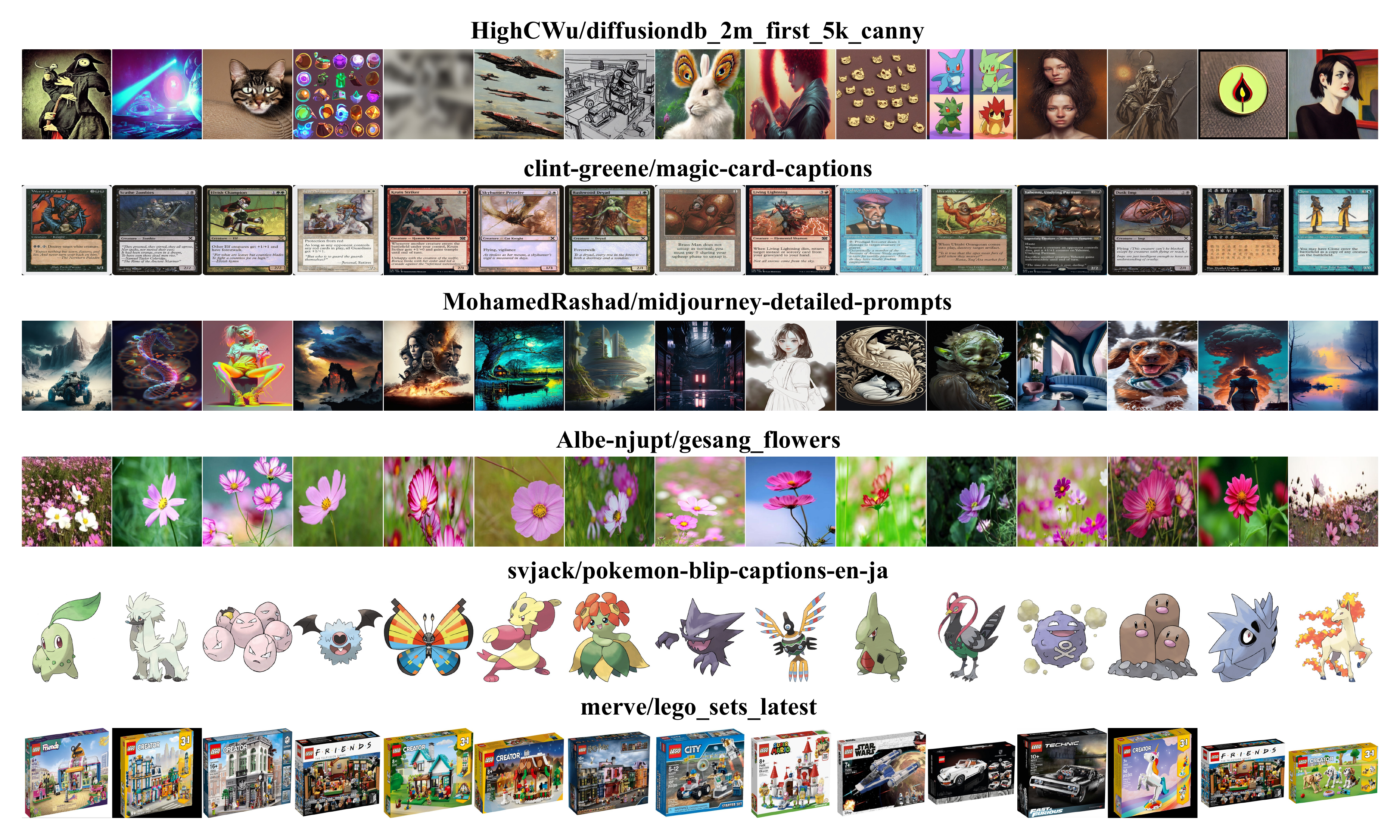}
    \caption{Examples of each dataset used in experiments.}
    \label{fig:dataset_examples}
\end{figure*}

\textbf{Dataset Examples.} Figure~\ref{fig:dataset_examples} showcases randomly selected examples from each dataset. For clarity, prompts are excluded from the visualizations. The original prompts can be accessed in the corresponding Hugging Face datasets.

\begin{table}[th!]
\centering
\caption{Average detection rate on different $\text{ef}_\text{construction}$, $M$ and $\text{ef}$ in HNSW implementation.}
\resizebox{.5\textwidth}{!}{%
\begin{tabular}{cc|c|cccccc}
\toprule
\midrule
\multirow{2}{*}{$\text{ef}$} & \multirow{2}{*}{Subset} & \multirow{2}{*}{$M$} & \multicolumn{6}{c}{$\text{ef}_\text{construction}$}\\
 &  &  & 50 & 100 & 200 & 300 & 400 & 500 \\\midrule
\multirow{15}{*}{200}           & \multirow{5}{*}{Flowers}     & 4                  & 0.8405 & 0.8349 & 0.8405 & 0.8405 & 0.8405 & 0.8405 \\
                                &                              & 8                  & 0.8410 & 0.8410 & 0.8415 & 0.8415 & 0.8415 & 0.8415 \\
                                &                              & 16                 & 0.8410 & 0.8415 & 0.8415 & 0.8415 & 0.8415 & 0.8415 \\
                                &                              & 32                 & 0.8410 & 0.8415 & 0.8415 & 0.8415 & 0.8415 & 0.8415 \\
                                &                              & 48                 & 0.8410 & 0.8415 & 0.8415 & 0.8415 & 0.8415 & 0.8415 \\\cline{2-9}
                                & \multirow{5}{*}{Lego Sets}   & 4                  & 0.2800 & 0.3035 & 0.3094 & 0.3094 & 0.3082 & 0.3082 \\
                                &                              & 8                  & 0.3082 & 0.3082 & 0.3094 & 0.3094 & 0.3094 & 0.3094 \\
                                &                              & 16                 & 0.3082 & 0.3082 & 0.3094 & 0.3082 & 0.3094 & 0.3094 \\
                                &                              & 32                 & 0.3082 & 0.3082 & 0.3094 & 0.3094 & 0.3094 & 0.3094 \\
                                &                              & 48                 & 0.3082 & 0.3082 & 0.3094 & 0.3094 & 0.3094 & 0.3094 \\\cline{2-9}
                                & \multirow{5}{*}{Magic Cards} & 4                  & 0.9770 & 0.9772 & 0.9772 & 0.9772 & 0.9772 & 0.9772 \\
                                &                              & 8                  & 0.9772 & 0.9772 & 0.9772 & 0.9771 & 0.9771 & 0.9771 \\
                                &                              & 16                 & 0.9771 & 0.9771 & 0.9771 & 0.9771 & 0.9771 & 0.9771 \\
                                &                              & 32                 & 0.9771 & 0.9771 & 0.9771 & 0.9771 & 0.9771 & 0.9771 \\
                                &                              & 48                 & 0.9771 & 0.9771 & 0.9771 & 0.9771 & 0.9771 & 0.9771 \\\midrule
\multirow{15}{*}{1000}          & \multirow{5}{*}{Flowers}     & 4                  & 0.8415 & 0.8415 & 0.8415 & 0.8415 & 0.8415 & 0.8415 \\
                                &                              & 8                  & 0.8415 & 0.8415 & 0.8415 & 0.8415 & 0.8415 & 0.8415 \\
                                &                              & 16                 & 0.8415 & 0.8415 & 0.8415 & 0.8415 & 0.8415 & 0.8415 \\
                                &                              & 32                 & 0.8415 & 0.8415 & 0.8415 & 0.8415 & 0.8415 & 0.8415 \\
                                &                              & 48                 & 0.8415 & 0.8415 & 0.8415 & 0.8415 & 0.8415 & 0.8415 \\\cline{2-9}
                                & \multirow{5}{*}{Lego Sets}   & 4                  & 0.2847 & 0.3035 & 0.3094 & 0.3094 & 0.3094 & 0.3094 \\
                                &                              & 8                  & 0.3094 & 0.3094 & 0.3094 & 0.3094 & 0.3094 & 0.3094 \\
                                &                              & 16                 & 0.3094 & 0.3094 & 0.3094 & 0.3094 & 0.3094 & 0.3094 \\
                                &                              & 32                 & 0.3094 & 0.3094 & 0.3094 & 0.3094 & 0.3094 & 0.3094 \\
                                &                              & 48                 & 0.3094 & 0.3094 & 0.3094 & 0.3094 & 0.3094 & 0.3094 \\\cline{2-9}
                                & \multirow{5}{*}{Magic Cards} & 4                  & 0.9770 & 0.9771 & 0.9771 & 0.9771 & 0.9771 & 0.9771 \\
                                &                              & 8                  & 0.9771 & 0.9771 & 0.9771 & 0.9771 & 0.9771 & 0.9771 \\
                                &                              & 16                 & 0.9771 & 0.9771 & 0.9771 & 0.9771 & 0.9771 & 0.9771 \\
                                &                              & 32                 & 0.9771 & 0.9771 & 0.9771 & 0.9771 & 0.9771 & 0.9771 \\
                                &                              & 48                 & 0.9771 & 0.9771 & 0.9771 & 0.9771 & 0.9771 & 0.9771
 \\
 \midrule
 \bottomrule
\end{tabular}%
}
\label{tbl:ablation_study}
\vspace{-.1in}
\end{table}

\begin{table*}[tbhp]
\resizebox{\textwidth}{!}{%
\begin{tabular}{lcccccccc}
\toprule
\midrule
Method & Learning Rate & Batch Size & \# Epochs & Image Size & LoRA Rank & LoRA Alpha & LoRA Target Layers & Precision \\\midrule
SD 1.4 (LoRA) & 0.001 & 64 & 150 & $512\times 512$ & 4 & 8 & {[}to\_k, to\_q, to\_v, to\_out.0{]} & float32 \\
SD 3 Medium (LoRA) & 0.001 & 64 & 150 & $512\times 512$ & 4 & 8 & {[}to\_k, to\_q, to\_v, to\_out.0{]} & float32 \\
SD 3 Medium (Full) & 0.0001 & 64 & 150 & $512\times 512$ & - & - & - & float32 \\
\midrule
\bottomrule
\end{tabular}%
}
\caption{Hyperparameter settings for model training.}
\label{tbl:hyperparameters}
\end{table*}

\section{Ablation Study}\label{apd:ablation}

To better understand the impact of key parameters on the performance of the HNSW implementation, we conducted an ablation study by varying the graph-related parameters \(M\) and \(\text{ef}\), as well as the construction parameter \(\text{ef}_\text{construction}\). Table~\ref{tbl:ablation_study} summarizes the average detection rates across three subsets: Flowers, Lego Sets, and Magic Cards, under a range of settings on SD 3 Medium with LoRA ($v=2^{12}$). 

The parameter \(M\) determines the maximum number of connections for each node in the graph. A larger \(M\) leads to denser graphs, which can improve accuracy at the cost of increased memory and computational overhead. The parameter \(\text{ef}_\text{construction}\) controls the size of the dynamic list of candidates during graph construction, influencing how exhaustive the neighborhood exploration is during index creation. Lastly, the query-time parameter \(\text{ef}\) defines the size of the candidate list used during the search operation, directly affecting the trade-off between accuracy and efficiency.

Across the three datasets, the Magic Cards consistently exhibited high detection rates, exceeding 97.7\% in all configurations, indicating that it is less sensitive to parameter tuning. In contrast, the Lego Sets showed significant variability. For \(\text{ef} = 200\), the detection rate improved notably with higher values of \(M\) (e.g., from 28\% at \(M=4\) to 30.82\% at \(M=8\) in $\text{ef} =200$ and $\text{ef}_\text{construction}=50$), but beyond \(\text{ef}_\text{construction}=100\), further increases in \(\text{ef}_\text{construction}\) provided diminishing returns. This suggests that while denser graphs and more exhaustive index construction improve accuracy for complex datasets, the benefits plateau at a certain point. For the Flowers, the detection rates remained stable at approximately 84.1\% across all parameter settings, indicating that this dataset is robust to variations in \(M\) and \(\text{ef}\).

\section{Additional Analysis}
\textbf{Training order and data augmentation.}
Our influence score is defined with respect to the final trained parameters $\theta$, rather than intermediate checkpoints. While training order affects the optimization trajectory, its effect is absorbed into the final parameter state used for influence estimation, especially under large-scale training with extensive shuffling. Data augmentation is treated as part of the training distribution; in practice, we report influence at the image level instead of treating each augmented view as a separate training sample.

\textbf{Timestep subsampling and gradient correlation.}
Adjacent diffusion timesteps are often highly correlated. \texttt{DMin} accounts for this by subsampling timesteps for influence estimation instead of computing gradients for all $T$ timesteps, as described in the main method.
This design reduces computation and storage while preserving performance.

\begin{figure}[t]
    \centering
    \vspace{-.15in}
    \includegraphics[width=0.95\linewidth]{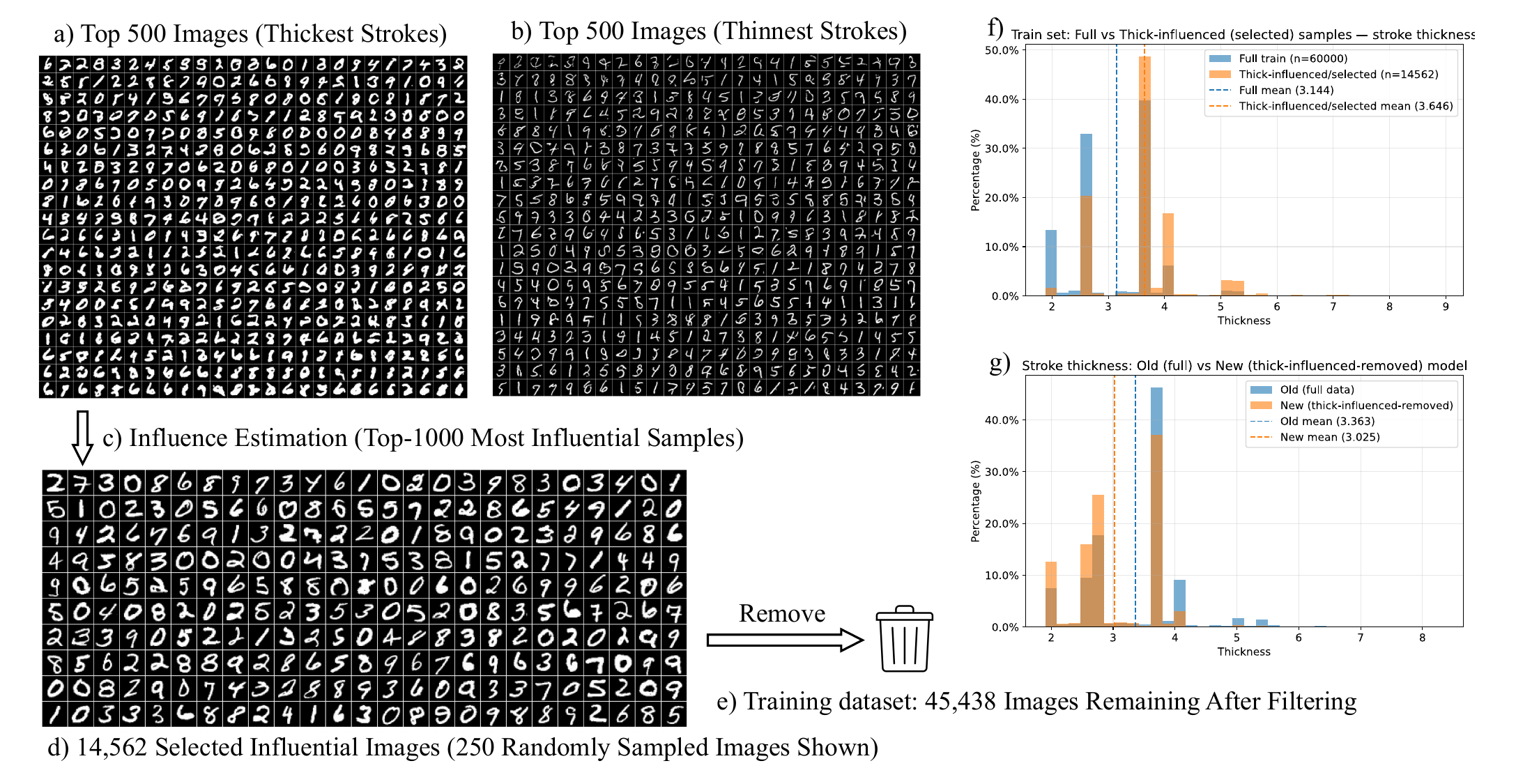}
    \vspace{-.12in}
    \caption{Removing highly influential MNIST training samples systematically shifts generated digits from thick to thinner strokes.}
    \label{fig:rebuttal_mnist}
    \vspace{-.1in}
\end{figure}

\textbf{Effect of removing influential samples.}
We further study whether removing highly influential samples changes model behavior on MNIST.
We first train an unconditional DDPM on the full 60,000-image training set and generate 5,000 samples.
Using a stroke-thickness metric, we select 500 generated images with the thickest strokes, run top-1000 influence estimation for each image, and deduplicate the retrieved training samples, yielding 14,562 influential samples.
After removing these samples and retraining on the remaining 45,438 images, generated digits shift toward thinner strokes, indicating that removing highly influential samples leads to systematic and interpretable changes in generation behavior (Figure~\ref{fig:rebuttal_mnist}).

\begin{figure}[t]
    \centering
    \includegraphics[width=0.95\linewidth]{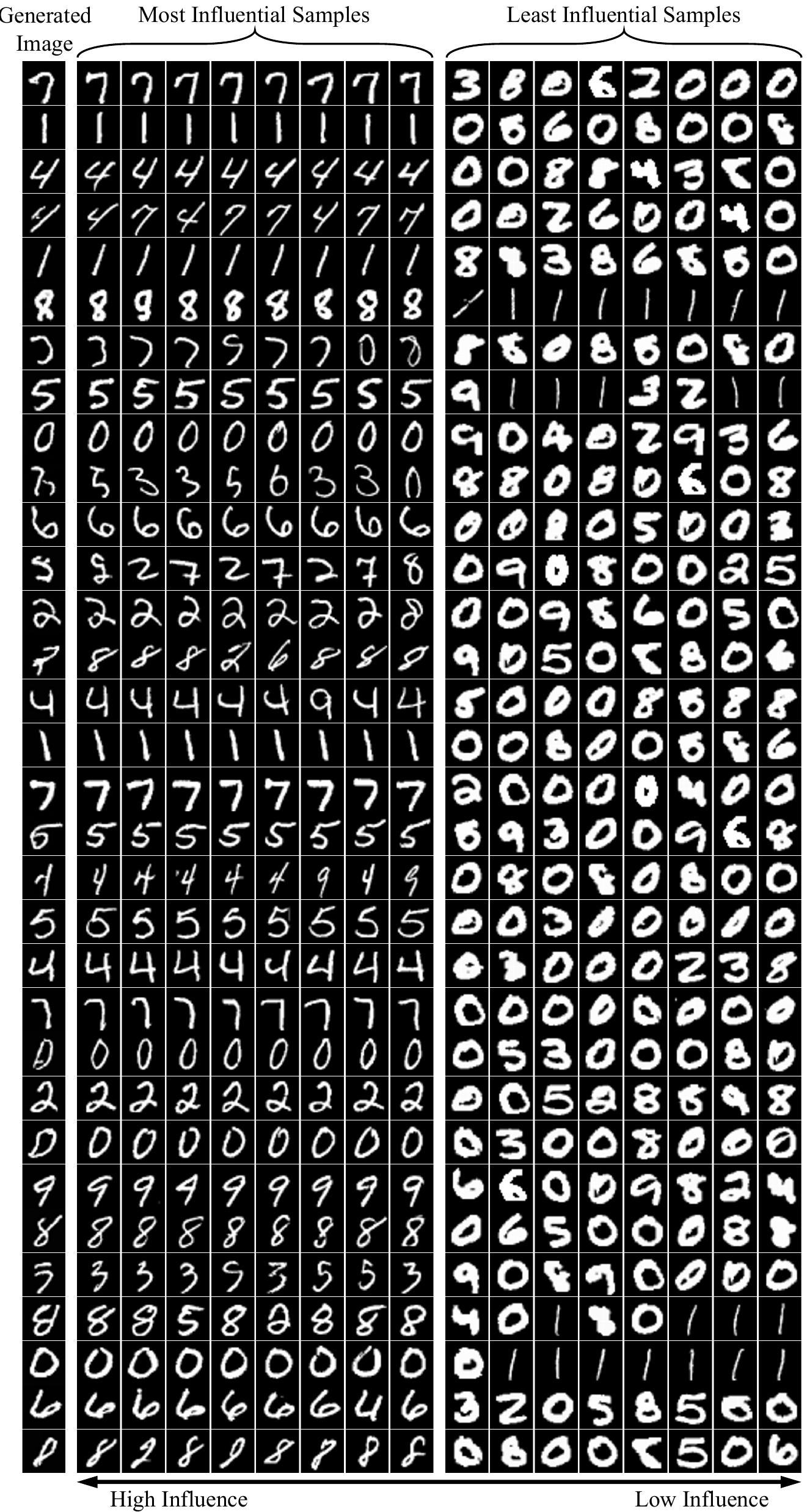}
    \caption{Additional visualizations for unconditional diffusion models on the MNIST dataset.}
    \label{fig:mnist_examples}
\end{figure}

\subsection{Baselines}\label{apd:baselines}

We compare the proposed \texttt{DMin} with seven baselines:
\begin{itemize}[itemsep=0pt, topsep=0pt, leftmargin=*, itemsep=0pt, parsep=0pt, partopsep=0pt]
\item \textbf{Random Selection:} Serves as a simple yet essential baseline where data points are selected randomly. This approach tests the performance against non-informed selection methods and ensures fairness in evaluation.
\item \textbf{SSIM:} A widely-used metric for assessing the similarity between two images or signals. This baseline tests the performance of similarity measures rooted in visual or structural fidelity.
\item \textbf{CLIP Similarity:} Exploits the feature embeddings generated by the CLIP, comparing their cosine similarity. It assesses how well general-purpose visual-language models can capture meaningful data relationships.
\item \textbf{LiSSA:} Measures the influence of training points on the model’s predictions by linearizing the loss function. This baseline provides a data-centric perspective on sample selection based on their impact on model training.
\item \textbf{DataInf:} Employs data influence techniques to prioritize training samples that most strongly influence specific predictions. It represents methods that utilize influence diagnostics in data selection.
\item \textbf{D-TRAK:} Focuses on tracking data’s training impact using gradient information. This baseline evaluates approaches that harness gradient dynamics for data importance measurement.
\item \textbf{Journey-TRAK:} Similar to D-TRAK but extends it to capture cumulative training effects over extended iterations. It benchmarks the ability of methods to consider long-term training trajectories in sample importance.
\end{itemize}

\section{Supplemental Visualization for Conditional Diffusion Models}\label{apd:vis_conditional}

\begin{figure*}
    \centering
    \includegraphics[width=\linewidth]{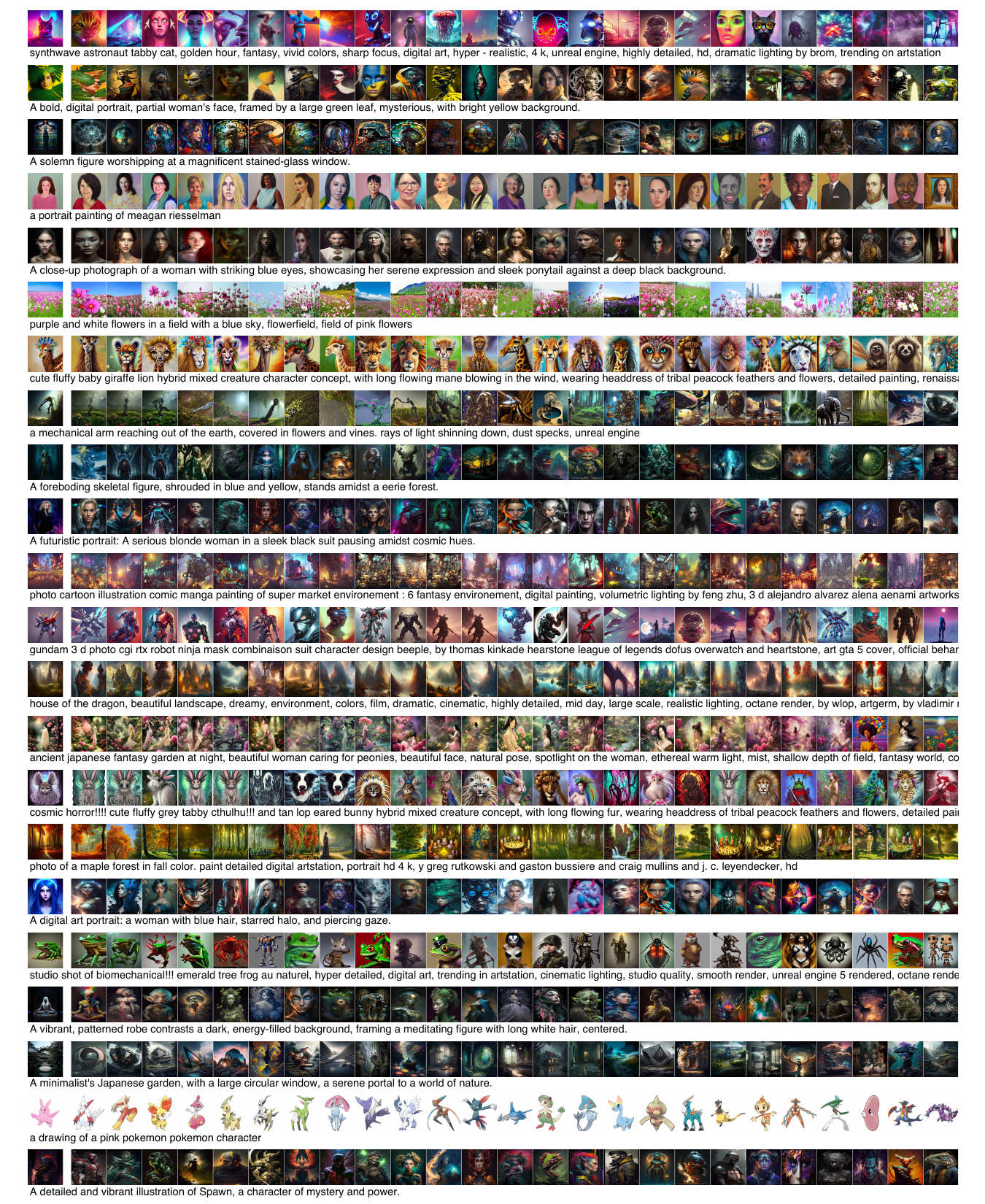}
    \caption{Examples of the top-25 most influential training data samples for the generated image (first column) on SD 3 Medium with LoRA, shown from high to low influence from left to right.}
    \label{fig:sd3_lora_examples}
\end{figure*}

We provide additional visualizations for unconditional models on the MNIST dataset in Figure~\ref{fig:mnist_examples} and for conditional models in Figure~\ref{fig:sd3_lora_examples}. Examples for other methods are omitted as they are nearly identical.

\end{document}